%% file: recordings_ds_arxiv.tex
\documentclass[10pt, a4paper]{article}
\pdfoutput=1
\usepackage{lrec}
\usepackage{multibib}
\usepackage{graphicx}
\usepackage{tabularx}
\usepackage{soul}
\usepackage{examples}
\usepackage[latin1]{inputenc}

\usepackage{url}
\usepackage{hyperref}

\usepackage{xstring}

\usepackage{verbatim}

\newenvironment{packed_item}{
	\begin{itemize}
		\setlength{\itemsep}{1pt}
		\setlength{\parskip}{0pt}
		\setlength{\parsep}{0pt}
	}{\end{itemize}}

\newtheorem{theorem}{Theorem}
\newtheorem{example}[theorem]{Example}

\usepackage{listings}
\title{A Recorded Debating Dataset}

\name{Shachar Mirkin$^1$, Michal Jacovi$^1$, Tamar Lavee$^{1,2}$, Hong-Kwang Kuo$^2$, Samuel Thomas$^2$, \\\textbf{\bf \large{Leslie Sager$^3$, Lili Kotlerman$^1$, Elad Venezian$^1$, Noam Slonim$^1$}}}

\address{$^1$IBM Research -- Haifa, Israel, $^2$IBM Watson -- Yorktown Heights, New York, USA, $^3$Tadiad, Tel Aviv, Israel\\
\{shacharm,michal.jacovi,lili.kotlerman,eladv,noams\}@il.ibm.com, tamar.lavee1@ibm.com,\\\{hkuo,sthomas\}@us.ibm.com,sagerleslie@gmail.com
}
		 
\abstract{
This paper describes an English audio and textual dataset of debating speeches, a unique resource for the growing research field of computational argumentation and debating technologies. We detail the process of speech recording by professional debaters, the transcription of the speeches with an Automatic Speech Recognition (ASR) system, their consequent automatic processing to produce a text that is more ``NLP-friendly'', and in parallel -- the manual transcription of the speeches in order to produce gold-standard ``reference'' transcripts. We release 60 speeches on various controversial topics, each in five formats corresponding to the different stages in the production of the data. The intention is to allow utilizing this resource for multiple research purposes, be it the addition of in-domain training data for a debate-specific ASR system, or applying argumentation mining on either noisy or clean debate transcripts. We intend to make further releases of this data in the future.
\thanks{This work is licensed under a CC BY-ND license.}
}

\begin{document}

\maketitleabstract

\input{intro}
\input{recording}
\input{asr}
\input{manual-trans}
\input{data}

\section{Acknowledgements}
We wish to thank the many speakers and transcribers who took part in the effort of creating this dataset. 

\section{References} \label{main:ref}
\bibliographystyle{lrec}
\bibliography{recordings_ds_arxiv}

\end{document}

%% file: intro.tex
\section{Introduction}\label{sec:intro}

Computational argumentation and debating technologies aim to automate the extraction, understanding and generation of argumentative discourse. This field has seen a surge in research in recent years, and involves a variety of tasks, over various domains, including legal, scientific writing and education.
Much of the focus is on argumentation mining, the detection of arguments in text and their classification~\cite{palau2009argumentation}, but many other tasks are being addressed as well, including argument stance classification~\cite{sobhani2015argumentation,bar2017improving}, the automatic generation of arguments~\cite{bilu2016claim}, identification of persuasive arguments~\cite{wei122016post}, quality assessment \cite{wachsmuth2017argumentation} and more. 
Multiple datasets are available for such research, mostly in English, such as the Internet Argument Corpus~\cite{WALKER12}, that consists of numerous annotated political discussions in internet forums, ArgRewrite~\cite{zhang-EtAl:2017:Long4}, a corpus of argumentative essay revisions, and the datasets released by IBM Research as part of the Debater Project~\cite{rinott2015show,aharoni2014benchmark}. \newcite{Lippi2016ArgumentationMS} list several additional such datasets. Further,~\newcite{wachsmuth2017building} have released an argument search engine over multiple debating websites, and~\newcite{aker2017projection} have initiated the projection of some datasets to languages other than English, such as Chinese.

All of the above are based on written texts, while datasets of spoken debates, outside of the political domain, are scarce. A spoken debate differs from a written essay or discussion not only in structure and content, but also in style as in any other case of spoken vs. written language. 
\newcite{Zhang2016ConversationalFI} made available transcripts from the Intelligence Squared\footnote{\url{http://www.intelligencesquaredus.org}} debating television show\footnote{\url{http://www.cs.cornell.edu/~cristian/debates/}}. The transcripts of the show are available on the show's site, and while they are of high quality, they do not match the audio recordings precisely, requiring substantial additional effort, if one wishes, for example, to use them as ASR training data.

With this paper we release a dataset of 60 audio speeches, recorded specifically for debating research purposes. We describe in detail the process of producing these speeches and their automatic and manual transcripts. This is a first batch of a larger set of recordings we intend to produce and release in the future.

%% file: recording.tex
\section{Recording the Speeches}

We recorded short speeches about debatable topics, with experienced speakers. This section describes the recording process. 

\paragraph{Recruiting and training the speakers}
Our team of speakers are all litigators or debaters, fluent or native English speakers, experienced in arguing about any given topic. 
The recruitment and training of the speakers included several steps. First, we interviewed potential speakers to evaluate their ability to argue about a topic when given only a short time to prepare. Then, we provided candidates with an essay to read aloud and record. Candidates were given technical guidelines to ensure high recording quality, including microphone configuration instructions and recording best-practices such as to record in a quiet environment, to use an external microphone and to maintain a fixed distance from the microphone while speaking. After listening to these recordings, we provided the speakers with feedback and repeated the process until the essay recordings were of good quality for the naked ear. 
Next, we provided each candidate with two motions (e.g. ``we should ban boxing'') and asked them to record a spontaneous speech supporting each motion, after a 10 minute preparation. 

All recordings -- three per candidate (one reading and two spontaneous speeches) -- were processed through automatic speech recognition and were sent to manual transcription, as described in the next sections. Comparing the automatic and manual transcripts, we computed the system's Word Error Rate (WER, the sum of substitution, deletion and insertion error rates) for each speech, and accepted candidates for whom the WER was below a pre-defined threshold of 10\%. That, to make sure that our ASR system is reasonably successful on their speeches.

\paragraph{The recording process}
All speakers received a list of motions, each with an ID and a short name (to be easily identified by human readers), and background information extracted from  Debatabase\footnote{\url{http://idebate.org/debatabase}} or Wikipedia. The speakers were directed to spend up to 10 minutes reviewing the motion's topic and preparing their arguments, and then immediately start recording themselves arguing in its favor for 4-8 minutes. The speakers were instructed not to search for further information about the topic beyond the provided description. 
The idea is to prevent multiple debaters who record a speech about the same topic from reaching the same resources (in particular debating websites), which may reduce the diversity of the ideas presented in the speeches. 
Example~\ref{ex:motion_info} shows a part of background information for the topic ``doping in sports''.

\begin{example}[\small{Topic background information}]\label{ex:motion_info} 
\begin{lstlisting}
At least as far back as Ben Johnson's steroid scandal at the 1988 Olympics, the use of performance-enhancing drugs in sports had entered the public psyche. Johnson's world record sprint, his win, and then, the stripping of his gold medal made news around the world. However, performance-enhancing drugs in sports do not begin with Johnson ...
\end{lstlisting}
\end{example}

%% file: asr.tex
\section{Automatic Speech Processing} \label{sec:asr}
Every recorded speech was automatically transcribed by a speaker-independent deep neural network ASR system. The system's acoustic model was trained on over 1000 hours of speech from various broadband speech corpora including broadcast news shows, TED talks\footnote{\url{https://www.ted.com/}} and Intelligence Squared debates\footnote{We semi-automatically aligned the transcripts and the audio, to overcome the inconsistency problem mentioned in Section~\ref{sec:intro}}. 
We used a $4$-gram language model with a vocabulary of 200K words, trained on several billion words that include transcripts of the above speech corpora and various written texts, such as news articles. 

The ASR system we used is similar to those described in~\cite{Soltau2013Neural,Soltau2014Joint}. We trained speaker-independent convolutional neural network (CNN) models on 40 dimensional log-mel spectra augmented with delta and double delta features. 
Each frame of speech is also appended with a context of 5 frames. The first CNN layer of the model has 512 nodes attached with $9 \times 9$ filters. Outputs from this layer are then processed by a second feature extraction layer, also with 512 nodes but using a set of $4 \times 3$ filters. The outputs from the second CNN layer are finally passed to 5 fully connected layers with 2048 hidden units each, to predict scores for 7K context-dependent states.
This speaker-independent ASR system performs on average at 8.4\% WER on the speeches we release with this paper.

\paragraph{}
Once a speech has been automatically transcribed, we obtain a text in the format shown in Example~\ref{ex:raw_asr}. Each token (including sentence boundary and silence markers \verb|<s>|, \verb|<s/>|, \verb|~SIL| ) is followed by the start and end time of its utterance, in seconds, relative to the beginning of the recording segment. 

This format is the basis for two versions of the data that we release for each speech: an automatically processed ``clean'' ASR version, and a manually transcribed one.
The steps for obtaining the former are described in Section~\ref{sec:asr-trans} The production of manual transcripts is described in Section~\ref{sec:manual-trans}

\begin{example}[\small{Raw ASR output}]\label{ex:raw_asr}
\begin{lstlisting}
<s>[0.000,0.660] we[0.660,0.830] should[0.830,1.060] allow[1.060,1.470] doping[1.470,2.010] in[2.010,2.240] sports[2.240,2.920] </s>[2.920,3.100] so[3.100,3.280] by[3.280,3.390] this[3.390,3.580] we[3.580,3.710] mean[3.710,4.110] </s>[4.110,4.140] steroids[4.140,4.950] </s>[4.950,5.080] human[5.080,5.390] growth[5.390,5.680] hormone[5.680,6.150] and[6.150,6.290] other[6.290,6.500] similar[6.500,6.960] drugs[6.960,7.490] ~SIL[7.490,7.670] should[7.670,7.890] be[7.890,8.000] allowed[8.000,8.330] in[8.330,8.430] pro[8.430,8.840] and[8.840,9.010] amateur[9.010,9.400] sports[9.400,10.030] </s>[10.030,10.070]
\end{lstlisting}
\end{example}

\subsection{ASR transcripts}\label{sec:asr-trans}
To obtain a ``clean'' version of the raw ASR output stream, we post-process it, as detailed below. After this processing, the text in Example~\ref{ex:raw_asr} is converted to the text in Example~\ref{ex:clean_asr}.

\begin{packed_item}
	\item \textbf{Removal of timing information}.
	\item \textbf{Removal of non-textual tokens:} Silence markers, \verb|~SIL|, appear whenever a relatively long pause has been detected in the speech; sentence boundary tags, \verb|<s>| and \verb|</s>|, denote predicted beginnings and ends of sentences. These are the result of the fact that the ASR language model was trained not only on spoken language transcripts, but also on written texts that contain punctuation marks. We have experimentally determined that, for our data, these predictions are not reliable enough to be utilized for sentence splitting on their own and used a dedicated method for this purpose, as described below. We also remove tags such as {\tt\%hes}, denoting unspecified speaker's hesitation, as well as other tokens denoting hesitation that were transcribed explicitly, such as {\tt{ah}}, {\tt{um}} or {\tt{uh}}.	
	\item \textbf{Abbreviations reformatting:} The ASR-produced underscored abbreviation (initialism) format (\verb|i_b_m|) is replaced with the standard all-caps one (\verb|IBM|).
	\item \textbf{Automatic punctuation and sentence splitting:} The automatically transcribed text contains no punctuation marks. 
	In downstream tasks, such as syntactic parsing, long texts are often difficult to handle, and we consequently split the stream of ASR output into sentences. Unlike typical sentence-splitting methods, whose main goal is to disambiguate between periods that mark end-of-sentence and those denoting abbreviations, here the text contains no periods, hence a different method is required. We employed a bidirectional Long Short-term Memory (LSTM) network \cite{Hochreiter:1997:LSTM} to predict commas and end-of-sentence periods over the ASR output. This neural network was trained on debate speeches, like the ones we share in this paper, and on TED talks, taken from the English side of the French-English parallel corpus from the IWSLT 2015 machine translation task \cite{wit3}.
\footnote{This is a simplified version of \cite{Pahuja:punct:2017}.}
	\item \textbf{Capitalization:} We apply basic truecasing to the text: capitalizing sentences' first letters and occurrences of ``I''. We have experimented with more sophisticated truecasing tools and abstained from employing them to the released texts due to mixed results.
\end{packed_item}

\begin{example}[\small{Clean ASR output}]\label{ex:clean_asr}
\begin{lstlisting}
We should allow doping in sports. 
So by this, we mean steroids, human growth hormone and other similar drugs should be allowed in pro and amateur sports.
\end{lstlisting}
\end{example}

%% file: manual-trans.tex
\section{Manual Transcription}\label{sec:manual-trans}
As mentioned, the ASR process produces imperfect texts. In order to obtain a ``reference'' text -- a precise transcript of the speech -- we employ human transcribers to post-edit the automatic transcript, i.e. correct its mistakes. 

\paragraph{Transcribers selection and training}
We invited 15 candidates to train as transcribers, all of which are native or fluent English speakers, experienced in linguistic annotation tasks. As a first test, we asked them to transcribe the same four speeches, after carefully reading the guidelines. We used their outcomes for creating ground-truth transcripts: for each speech, we compared its transcripts pair-wise, listened carefully to points of differences, and created a ``gold-transcript'' that resolved all differences between the individual transcripts. Using these four gold-transcripts, we scored the work of the individual transcribers, and accepted as transcribers nine of the candidates whose transcripts were at least 98\% accurate. 
They were further trained by transcribing ten speeches each, and getting feedback on them upon our review. Once done, we considered them ``experienced transcribers''.

\paragraph{Transcription methodology}
In our experience, starting from initial transcripts produced by ASR can halve the time necessary to produce reference transcripts, while maintaining similar transcript quality.  This is particularly true if the ASR is highly accurate since it reduces the number of corrections the human transcriber has to make. 
One should be aware, however, that this procedure can introduce bias, depending on how conscientious the human transcriber is.  An inexperienced or less conscientious transcriber may neglect to correct some ASR mistakes. 

It is also easier for human transcribers to process shorter segments of speech, especially if they have to
listen multiple times to unclear segments. Hence, to speed up the process of human transcription, the audio and transcript
are first segmented by cutting them at silences longer than 500ms.  Excessively long audio segments are then further divided
at their longest silences, which must be at least 100ms.
Note that the resulting segments do not necessarily correspond to linguistic boundaries or to where punctuation marks should be placed. Instead, in spontaneous speech, a person may pause in the middle of a sentence when faced with an increased cognitive load, e.g. when trying to recall a word.
Similar methods of using ASR output as a basis for manual transcription were applied, e.g., by \cite{Park-Zeanah:2005} and \cite{matheson2007voice}, for the purpose of transcribing interviews for interview-based research.

The human transcribers used Transcriber\footnote{\url{http://trans.sourceforge.net/en/presentation.php}; We used version 1.5.1}, a tool for assisting manual annotation of speech signals through a graphical user interface. 
The tool synchronizes the text with the audio, and allows the human transcriber to review the text while listening to the audio, and easily pause, fix, annotate, and continue listening from a selected segment. 

On average, the time needed for manual transcription by experienced transcribers was approximately five times the duration of the audio file.
An example of the input to the tool -- the output of the above-mentioned segmentation process -- is presented in Example~\ref{ex:trs:in}. The output of the post-edition, which uses the same format, is shown in Example~\ref{ex:trs:out}.

The guidelines used for manual transcription explain how to deal with cases such as speaker hesitation, repetitions and utterance of incomplete words, what punctuation marks to use\footnote{The ASR does not produce punctuation marks; it turned out that the transcribers preferred adding them, as it made the text more readable. Punctuation also makes the texts more accessible for analysis and annotation and may be helpful for some automatic processing tasks.}, how to write abbreviations, numbers, etc. The main principles are that the transcripts should be accurate with respect to the source, capture as much signal as possible, and that they should maintain a uniform format that can be easily parsed in subsequent processing.\footnote{The transcription guidelines are shared with the released data.} 

\begin{example}[\small{Input for manual transcription}]\label{ex:trs:in}
\begin{lstlisting}
<Sync time="18.020"/>
doping is the use of performance enhancing drugs
<Sync time="21.290"/>
at what i
<Sync time="22.030"/>
am talking about sports i am of course referring to
<Sync time="25.015"/>
a competitive sports 
<Sync time="26.630"/>
for example the olympics
<Sync time="28.320"/>
or other kinds of competitions
<Sync time="30.040"/>
like a true the fonts 
<Sync time="31.800"/>
and etcetera
\end{lstlisting}
\end{example}

\begin{example}[\small{Output of manual transcription}]\label{ex:trs:out}
\begin{lstlisting}
<Sync time="18.020"/>
doping is the use of performance enhancing drugs .
<Sync time="21.290"/>
uh when i
<Sync time="22.030"/>
am talking about sports i am of course referring to
<Sync time="25.015"/>
uh competitive sports ,
<Sync time="26.630"/>
for example the olympics
<Sync time="28.320"/>
or other kinds of competitions
<Sync time="30.040"/>
like uh tour de france
<Sync time="31.800"/>
uh etcetera ,
\end{lstlisting}
\end{example}

\subsection{Reference Transcripts}
Some of the annotations in the post-edited transcripts are mostly useful for ASR training, as in the case of word mispronunciation and its correction (e.g. ``{lifes/lives}''), while others contain signals that may also be useful for downstream text processing. 

Our approach in producing the reference transcripts was to remove all non-textual annotations, producing a text-only version of the transcription, that can be used as-is, e.g. for argument extraction. From the Transcriber's output, we first removed all SGML tags and merged the lines into a single stream. We then removed incomplete words and mispronounced words (replacing them with the correct pronunciation); similarly to the raw ASR post-processing, we removed annotations, hesitations, reformatted abbreviations and applied basic truecasing. Then, we detokenized the text, i.e. removed any unnecessary spaces between tokens, for example, before a punctuation mark. Lastly, we applied automatic spell-checking to detect typos and formatting errors, and sent the identified instances of possible typos for review. Example~\ref{ex:clean_ref} shows the text segment from Example~\ref{ex:trs:out} after going through this cleaning. 

\begin{example}[\small{Clean reference transcript}]\label{ex:clean_ref}
\begin{lstlisting}
Doping is the use of performance enhancing drugs.
When I am talking about sports I am of course referring to competitive sports, for example the olympics or other kinds of competitions like tour de france etcetera,
\end{lstlisting}
\end{example}

%% file: data.tex
\section{Dataset}\label{sec:data}

\begin{table}
	\centering
	{\small
		\begin{tabular}{|l|l|}
		\hline
			\textbf{Extension} & \textbf{Description}\\
		\hline
		
			\tt{wav} & Recorded speeches \\
			\tt{asr} & Raw automatic transcripts \\
			\tt{asr.txt} & Clean automatic transcripts \\
			\tt{trs} & Manual transcripts \\  
			\tt{trs.txt} & Clean manual transcripts (references)\\
 			\hline
		\end{tabular}
		\caption{Summary of the dataset file types.}\label{tab:files}
		}
\end{table}

\begin{table}[!h]
\begin{center}
\begin{tabularx}{\columnwidth}{|c|X|c|c|}
		\hline	
			 \textbf{ID} & \textbf{Topic} & \textbf{Speeches} & \textbf{WER (\%)} \\		
			\hline
		1   & Violent video games  & 6 & 7.4 \\ 
		21  & One-child policy  & 5 & 8.3 \\ 
		61  & Doping in sports  & 5 & 7.7 \\ 
		101 & Affirmative action	& 5 & 9.6  \\
		121 &	Boxing & 5 & 9.6  \\ 
		181 &	Multiculturalism  & 2 & 8.5  \\ 
		381 &	The monarchy  & 5 & 7.3  \\ 
		482 & Cultivation of tobacco & 3 & 8.2 \\ 
		483 &	Freedom of speech  & 5 & 6.7 \\ 
		602 & School vouchers	  & 5 & 7.2  \\ 
		644 &	Year-round schooling & 1 & 8.9 \\	
		681 & Intellectual property & 3 &10.9 \\ 
		701 & Endangered species & 2 & 6.8 \\ 
		841 & Blasphemy & 3 &9.3\\ 
		881 & Holocaust denial & 3 & 9.8\\ 
		945 & Infant circumcision & 2 & 11.2\\ 
		\hline
 \end{tabularx}
\caption{List of motion topics in our dataset, and the number of speeches per topic. The right column shows the average WER across speeches of the topic, when using the speaker-independent ASR model.}\label{tab:data}
\end{center}
\end{table}

The dataset we created was generated through the process described in the previous sections. We release all file types, including raw and clean versions, to enable research based on various signals, including audio-based ones, such as prosody or speech rate, and to allow performing different post-processing. Table~\ref{tab:files} summarizes the files that are obtained and released for each debatable topic.

The dataset we release includes 60 speeches for 16 motions from \cite{rinott2015show}, recorded by 10 different speakers.\footnote{Currently, the list contains only a single female speaker; we are making an effort to recruit more female debaters.}
Table~\ref{tab:data} provides details about the recordings included in the dataset. 

There is a large variance in WER across different debate recordings, and between different speakers. The WER of any specific debate can vary depending on the degree of mismatch with the ASR acoustic and language models. Examples of mismatch include differences in speaker voice, speaking style and rate, audio capture (microphone type and placement), ambient noise, word choice and phrasing, etc. By reducing mismatch through model adaptation of speaker-dependent acoustic models, the WER can be significantly reduced. For instance, with adaptation using about 15 minutes of a speaker's data, WER of a speech from topic 61 was reduced from 12.9\% to 8.6\%, and of a speech from topic 483, from 12.2\% to 9.7\%.

The dataset is freely available for research at \url{https://www.research.ibm.com/haifa/dept/vst/mlta_data.shtml}.